\documentclass[runningheads]{llncs}

\usepackage{eccv}

\usepackage{eccvabbrv}

\usepackage{graphicx}
\usepackage{booktabs}
\usepackage{caption}
\usepackage{enumitem}
\usepackage{amsmath}
\usepackage{tabularx} 
\usepackage{makecell}
\usepackage{marvosym}

\newcommand \footnoteONLYtext[1]{
	\let \mybackup \thefootnote
	\let \thefootnote \relax
	\footnotetext{#1}
	\let \thefootnote \mybackup
	\let \mybackup \imareallyundefinedcommand
}

\usepackage{hyperref}

\begin{document}
\title{CVT-Occ: Cost Volume Temporal Fusion \\for 3D Occupancy Prediction} 


\author{Zhangchen Ye \textsuperscript{1*} \and
Tao Jiang \textsuperscript{1,2*} \and
Chenfeng Xu\textsuperscript{3} \and
Yiming Li\textsuperscript{4} \and \\
Hang Zhao\textsuperscript{1,2,5\Letter}
}

\footnoteONLYtext{*: Equal contribution.}
\footnoteONLYtext{\Letter:  Corresponding to: hangzhao@mail.tsinghua.edu.cn}

\authorrunning{Z. Ye et al.}

\institute{
    \textsuperscript{1}IIIS, Tsinghua University \hspace{1em}
    \textsuperscript{2}Shanghai AI Lab \hspace{1em}
    \textsuperscript{3}UC Berkeley \\
    \textsuperscript{4}New York University \hspace{1em}
    \textsuperscript{5}Shanghai Qi Zhi Institute
}

\maketitle
\begin{abstract}
Vision-based 3D occupancy prediction is significantly challenged by the inherent limitations of monocular vision in depth estimation. This paper introduces CVT-Occ, a novel approach that leverages temporal fusion through the geometric correspondence of voxels over time to improve the accuracy of 3D occupancy predictions. By sampling points along the line of sight of each voxel and integrating the features of these points from historical frames, we construct a cost volume feature map that refines current volume features for improved prediction outcomes. Our method takes advantage of parallax cues from historical observations and employs a data-driven approach to learn the cost volume. We validate the effectiveness of CVT-Occ through rigorous experiments on the Occ3D-Waymo dataset, where it outperforms state-of-the-art methods in 3D occupancy prediction with minimal additional computational cost. The code is released at \url{https://github.com/Tsinghua-MARS-Lab/CVT-Occ}. 
\end{abstract}
\section{Introduction}
\label{sec:intro}
Vision-based 3D semantic occupancy prediction is rapidly evolving in the domain of 3D perception, driven by its critical applications in autonomous driving, robotics, and augmented reality. The task aims to estimate the occupancy state and semantic label of every voxel within a 3D space from visual inputs~\cite{tpvformer,tian2023occ3dlargescale3doccupancy, zhang2023occformer, sima2023sceneoccupancy}.

Despite its critical importance, 3D occupancy prediction presents significant challenges. When relying solely on monocular vision, these challenges are particularly pronounced due to the inherent ambiguity in estimating depth from a single image. While stereo vision has been proposed as a solution to augment depth estimation accuracy~\cite{li2023stereovoxelnet}, its application remains limited in practice. It is impractical to use stereo cameras for widespread applications in autonomous vehicles and robotic systems because of the requirement for extensive calibration and re-calibration. An alternative and more promising approach is the employment of multi-view temporal fusion, which can utilize the extended multi-view baselines available over time to enhance 3D perception tasks.

\begin{figure}[!t]
  \centering
  \includegraphics[width=\linewidth]{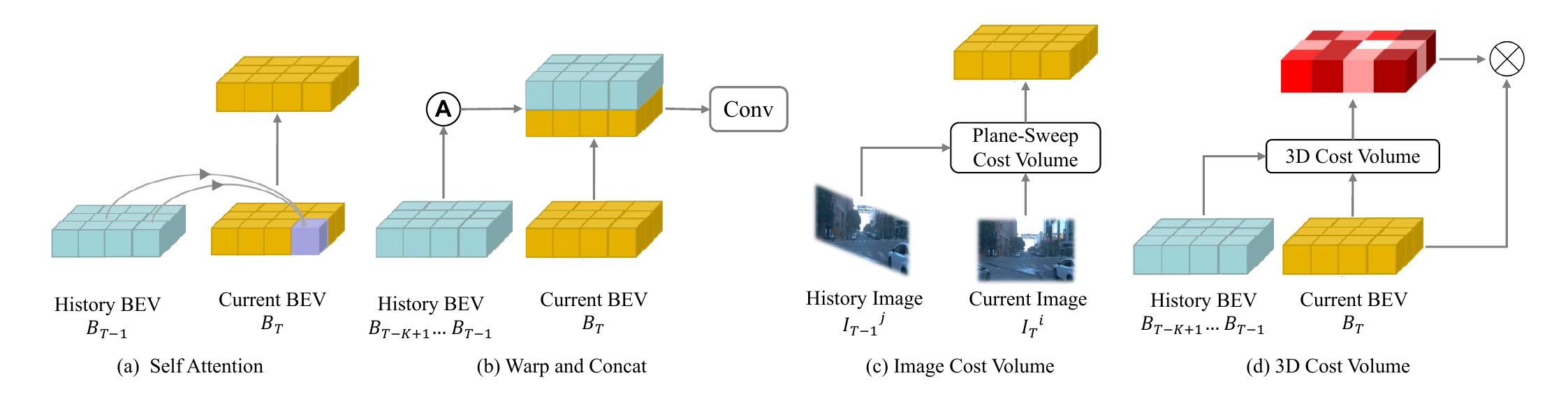}
  \caption{\textbf{Comparison of Temporal Fusion Methods.} Illustrated are four key approaches: (1) Temporal Self-Attention~\cite{li2022bevformer}, leveraging attention mechanisms for temporal integration; (2) Warp and Concat~\cite{huang2022bevdet4d, yang2022bevformerv2, wang2023panooccunifiedoccupancyrepresentation}, combining features across frames and fusing them through convolution; (3) Cost Volume Construction in image space~\cite{solofusion}, constructing cost volume from image input of different frames and leveraging plane-sweep volumes for depth map generation; and (4) Our Proposed Method, which involves constructing a temporal cost volume in 3D space to enhance feature refinement. In the figure, \textcircled{\tiny{A}} and $\otimes$ represent coordinate alignment and element-wise product, accordingly.}
  \label{fig:teaser}
\end{figure}

Recent advancements in visual 3D object detection have shown the potential of incorporating temporal observations to bolster detection performance~\cite{li2022bevformer, huang2022bevdet4d, yang2022bevformerv2, wang2023panooccunifiedoccupancyrepresentation, solofusion}. In our research, we have identified and categorized the emerging temporal fusion methods into three paradigms, as depicted in Figure~\ref{fig:teaser}. 

The first two paradigms are categorized as warp-based methods. These methodologies primarily involve the alignment of Bird's Eye View (BEV) feature maps across multiple temporal instances, utilizing relative camera poses obtained from Inertial Measurement Units (IMUs). Subsequently, the aligned features undergo integration using two primary approaches. The first approach employs self-attention mechanisms, as depicted in Figure~\ref{fig:teaser}(a). A representative method utilizing this approach is BEVFormer~\cite{li2022bevformer}. Alternatively, the second approach integrates aligned features through concatenation, as illustrated in Figure~\ref{fig:teaser}(b). Representative methods employing this approach include BEVDet4D~\cite{huang2022bevdet4d},  BEVFormerv2~\cite{yang2022bevformerv2}, and PanoOcc~\cite{wang2023panooccunifiedoccupancyrepresentation}. These methods emphasize feature fusion using concatenation followed by convolutional operations. Despite their potential, these methods primarily leverage temporal information in an implicit manner, lacking a robust understanding of temporal geometry. As a result, they fall short in fully exploiting the geometric constraints inherent in 3D space. 

In contrast, cost volume-based methods, exemplified by (c) in Figure~\ref{fig:teaser} and represented by SOLOFusion~\cite{solofusion}, draw inspiration from stereo matching techniques. Constructing a cost volume from images captured from different viewpoints in the temporal sequence enables the utilization of geometric constraints to obtain depth-aware features. However, for multi-view vision tasks, the sheer number of image pairs can lead to significant computational overhead when fusing longer time span temporal information.

Recognizing the limitations of these approaches, we propose a novel paradigm, as illustrated in (d) in Figure~\ref{fig:teaser}. In this paper we introduce CVT-Occ, an innovative temporal fusion method designed to capitalize on the geometric correspondences of voxels over time, thereby enhancing occupancy prediction accuracy. Our approach involves sampling points along the line of sight for each voxel - defined as the line connecting the voxel to the optical center of the camera - and identifying the corresponding 3D locations of these points in historical frames based on relative camera poses. We then sample the features of these points in historical frames and integrate them with the current voxel features to construct a cost volume feature map. This map is subsequently utilized to refine the features of the current volume, thereby improving occupancy prediction. 

Compared with paradigms (a) and (b) in Figure~\ref{fig:teaser}, our proposed CVT-Occ explicitly utilizes parallax cues to refine the depth of 3D voxel. In contrast to paradigm (c), our approach avoids per-image-pair cost volume computation, achieving superior performance with minimal additional computational overhead. CVT-Occ distinguishes itself by leveraging the inherent parallax information in historical observations and employing a data-driven approach to learning the cost volume. Our experiments on the Occ3D-Waymo dataset demonstrate that CVT-Occ achieves state-of-the-art 3D occupancy prediction performance. 


\section{Related Work}
\subsection{3D Occupancy Prediction}
The goal of 3D occupancy prediction is to estimate the occupancy of each quantized voxel in 3D space. This task is originally rooted in Occupancy Grid Mapping (OGM)~\cite{Thrun_2002, Moravec_Elfes_2005} for mobile robot navigation, where the robot is equipped with range sensors (e.g., LiDAR) and navigates in static environments. Recent works transit to a more generic scenario: they make the occupancy prediction with visual systems in dynamic environments. 
MonoScene~\cite{Monoscene} reconstructs the 3D scene from sparse prediction with LiDAR points. VoxFormer ~\cite{VoxFormer} makes dense voxel predictions leveraging semantic labels and depth estimation from monocular RGB images. TPVFormer~\cite{tpvformer} proposes a tri-perspective view decomposition method for efficient 3D occupancy prediction. OccFormer~\cite{zhang2023occformer} presents a dual-path transformer network to effectively process the 3D volume for semantic occupancy prediction.
Besides, these advancements have led to new benchmarks such as visual 3D occupancy prediction~\cite{wang2023openoccupancylargescalebenchmark, tian2023occ3dlargescale3doccupancy, sima2023sceneoccupancy} and semantic scene completion~\cite{li2023sscbench}. We emphasize that our proposed method is a simple and plug-and-play module that seamlessly integrates with existing occupancy prediction pipelines, significantly improving their performance.
\subsection{Temporal Fusion for 3D Perception}
Leveraging temporal information for occupancy prediction is a natural strategy since it provides sufficient spatial information for building geometric representations \cite{solofusion,xu2023nerfdet}. A common practice of temporal fusion in visual 3D object detection is BEV feature alignment. It fuses temporal information by warping past BEV features to the current time according to camera view transformation from relative camera poses at different time steps~\cite{li2022bevformer,huang2022bevdet4d}. It has been shown that short-term frames can significantly help improve the 3D detection performance. However, we find that minor improvements are brought up when more frames come in with simply warping and fusion. Recent work UniFusion~\cite{qin2023unifusionunifiedmultiviewfusion} proposes a new long temporal fusion method. It creates ``virtual views'' for the temporal features as if they are present in the current time and directly access all useful history features. However, this method suffers from intensive computation costs when the number of frames increases. SOLOFusion~\cite{solofusion} is a more efficient and effective model that balances the resolution and time stamps, and transforms the temporal fusion problem into a temporal multi-view stereo problem.
\subsection{Stereo Matching and Multi-View Stereo}
Stereo matching is rooted in constructing 3D cost volumes from 2D image features ~\cite{Chang_Chen_2018PyramidStereoMatchingNetwork, GANet, AnytimeStereoImageDepthEstimationonMobileDevices} and predicting the depth map from the cost volume. Recent works further advance it by proposing correlation-based cost volumes like GCNet~\cite{cao2019gcnet}. Meanwhile, multi-view stereo methods(~\cite{DeepMVS, yao2018mvsnetdepthinferenceunstructured, MVDepthNet, im2019dpsnetendtoenddeepplane, cheng2020deepstereousingadaptive, Gu_2020_CVPRCascadeCostVolum}) leverage plane-sweep volumes for depth map generation. A recent trend of using stereo matching~\cite{chen2020dsgn, li2023bevstereo, li2023bevstereo++} focused on 3D perception in autonomous driving. While existing work in stereo matching often utilizes stereo image pairs, this format is unsuitable for handling multi-view and multi-temporal inputs in autonomous driving scenarios. Additionally, these approaches require generating a Plane-Sweep Volume for each image pair, leading to inefficiencies. Furthermore, recent advancements like OccDepth~\cite{OccDepth} demonstrate the potential of implicitly learning correlations between stereo images to improve 3D depth-aware feature fusion. Our proposed Temporal Cost Volume method, designed specifically for multi-view and multi-temporal data, offers a more effective and efficient solution. The core innovation of the CVT module lies in its ability to construct a comprehensive 3D representation of the scene by aggregating information over time. 
\section{Methodology}
\subsection{Problem Setup}
Given only RGB images as inputs, the model is designed to predict a dense semantic scene within a specified volume. Specifically, we utilize as input the current timestamp \(t\) along with previous images, denoted by \( \mathbf{I} = \{ \mathbf{I}_t^i, \mathbf{I}_{t-1}^i, \ldots \}_{i=1}^{L} \), where \(L\) represents the number of cameras. Our output is defined as a voxel grid \(Y_t \in \{c_0, c_1, \ldots, c_M\}^{H \times W \times Z}\), situated in the coordinate frame of the ego-vehicle at timestamp \(t\). Each voxel within this grid is either empty (indicated by \(c_0\)) or occupied by a specific semantic class among \(\{c_1, c_2, \ldots, c_M\}\). Here, \(M\) denotes the total number of classes of interest, while \(H\), \(W\), and \(Z\) represent the length, width, and height of the grid, respectively. The primary goal is to train a neural network \(\Theta(\cdot)\) to produce a semantic voxel grid \(Y_t = \Theta(\mathbf{I})\) that approximates the ground truth \(\hat{Y}_t\). It is important to note that the model must learn both 3D geometry and semantics solely from vision-based inputs, without the aid of depth measurements from LiDAR. Inferring 3D occupancy from 2D images presents a significant challenge and necessitates the design of efficient methods to leverage geometric constraints for learning accurate geometry.

\begin{figure}[t!]
  \centering
  \includegraphics[width=\linewidth]{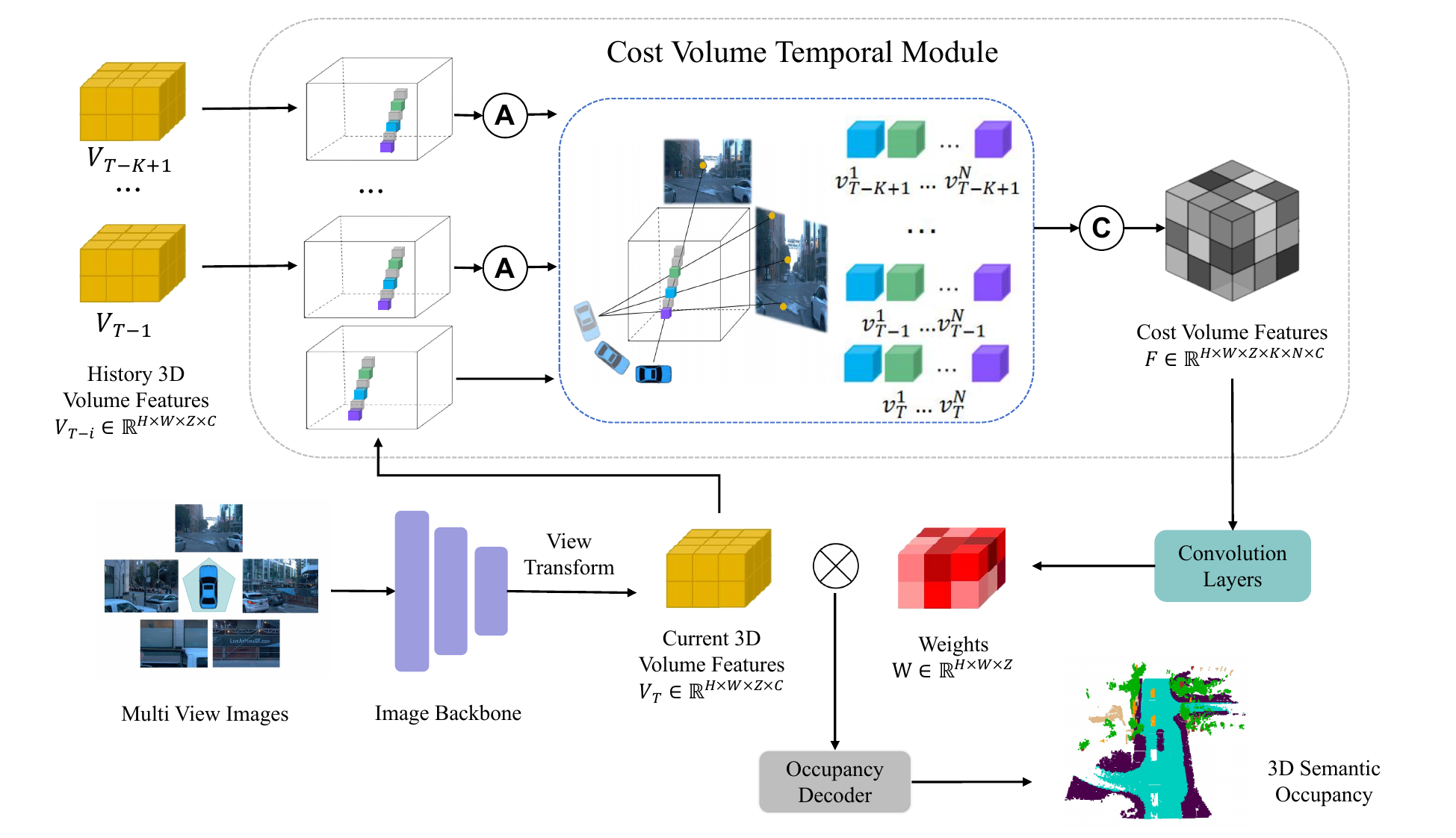}
  \caption{\textbf{Overall Architecture of CVT-Occ. }The image backbone extracts multi-scale features from multi-view images, which are transformed into 3D volume features denoted as $\mathbf{V} \in \mathbb{R}^{H \times W \times Z \times C}$. The Cost Volume Temporal Module samples points along the line of sight within the current volume and projects them onto $K-1$ historical frames, resulting in $K \times N$ 3D volume features. These features are concatenated to construct cost volume features $\mathbf{F} \in \mathbb{R}^{H \times W \times Z \times (K\times N) \times C}$. Convolution layers are then applied to generate weights $\mathbf{W} \in \mathbb{R}^{H \times W \times Z}$, refining the depth of 3D voxel. Finally, an occupancy decoder produces 3D semantic occupancy predictions. In the figure, \textcircled{\tiny{A}}, \textcircled{\tiny{C}}, and $\otimes$ represent coordinate alignment, concatenation, and element-wise product, respectively.}
  \label{fig:network}
\end{figure}
\subsection{Overall Architecture}
In this section, we depict the comprehensive architecture of CVT-Occ, depicted in Figure~\ref{fig:network}. Our framework processes multi-frame, multi-view images to first extract multi-scale features through an image backbone. Subsequently, these features from image space are transformed into BEV space features, which are refined by a BEV encoder to generate 3D volumetric representations. There is a rich corpus of research focused on the transition from image space to BEV features. One line of works follows the lifting paradigm proposed in LSS~\cite{LSS}; they explicitly predict a depth map and lift multi-view image features onto the BEV plane~\cite{huang2021bevdet,huang2022bevdet4d,li2023bevdepth,li2023bevstereo,solofusion}. Another line of works inherits the spirit of querying from 3D to 2D in DETR3D~\cite{wang2022detr3d}; they employ learnable queries to extract information from image features by cross-attention mechanism~\cite{li2022bevformer,jiang2023polarformer,liu2022petrv2}. It is crucial to underscore that our proposed Cost Volume Temporal (CVT) module is versatile and compatible across different strategies for feature transformation from image to volume space. For our experimental validation, we employ BEVFormer~\cite{li2022bevformer} as the foundation to generate 3D volume features. These features are subsequently refined and enhanced through the CVT module, demonstrating the efficacy and adaptability of our approach in leveraging temporal and spatial dynamics for enriched 3D scene comprehension. Finally an occupancy decoder is used to generate final voxel prediction results.
\subsection{Cost Volume Temporal Module}
Since direct depth information is not available for each pixel, transforming image features into 3D space introduces ambiguity; for instance, a single pixel may correspond to multiple voxels along the line of sight. To address this challenge, we propose the Cost Volume Temporal module, which leverages temporal data to infer depth information and resolve this ambiguity. Specifically, we construct 3D cost volume features using both historical and current BEV features. These cost volume features are then used to derive learned weights, which are subsequently employed to refine the current BEV features.

\noindent\textbf{3D Volume Features.} We pre-define the volume \( \mathbf{V} \in \mathbb{R}^{H \times W \times Z} \), where \( H, W, Z \) represent the spatial dimensions of the volume space. Each voxel in the volume space corresponds to a cubic space in the real world with a size of \(s\) meters. By default, the center of the volume corresponds to the position of the ego car. For each voxel \( v = (i, j, k) \), its corresponding 3D location \( p = (x, y, z) \) can be calculated by Eqn.~\ref{eq:voxel2location}:
\begin{equation}
\label{eq:voxel2location}
x = \left( i - \frac{W}{2} \right) \times s, \quad y = \left( j - \frac{H}{2} \right) \times s, \quad z = \left( k - \frac{Z}{2} \right) \times s
\end{equation}
The BEV feature is $\mathbf{F}^{\text{bev}}_{t} \in \mathbb{R}^{H\times W \times E}$, where $t$ represents the timestamp and $E$ is the embedding dimensions. We reshape the BEV feature to volume space as the input of the CVT module, producing $\mathbf{V}_t \in \mathbb{R}^{H \times W \times Z \times C}$. 

\noindent\textbf{Construct Cost Volume Feature.} Given the ambiguity along the line of sight, for each point \( \mathbf{p_t} \), we sample several additional points \( \{\mathbf{p_t^i}\}_{i=1}^{N} \) within the current volume. Specifically, we calculate the sight direction \( \mathbf{d_t} \in \mathbb{R}^{3} \), which is the vector from the center of the volume to the point \( \mathbf{p_t} \), and then sample points using specific strides \( \{n^i\}_{i=1}^{N} \) as shown in Eqn.~\ref{eq:sample}:
\begin{equation}
\label{eq:sample}
\mathbf{p_t^i} = \mathbf{p_t} + \mathbf{d_t} \times \mathbf{n^i}
\end{equation}
Since these points project to the same pixel in the image space, points along the same line tend to have similar features. To accurately distinguish the correct position corresponding to the pixel, we use historical BEV features to obtain complementary information. Here comes the core insight of our proposed Cost Volume Temporal Module. Projecting these points into the historical coordinate frame ensures they are no longer in the same line of sight. This parallax provides additional information from the historical BEV features, which helps to reduce depth ambiguity in the current frame. Projection matrix \( \mathbf{P_t} \in \mathbb{R}^{4 \times 4} \) can transform points from the coordinate frame of ego-vehicle to the global coordinate frame. Therefore the points \( \{\mathbf{p_t^i}\}_{i=1}^{N} \) are projected onto \( K-1 \) historical frames through the projection matrix as described in Eqn.~\ref{eq:project}:
\begin{equation}
\label{eq:project}
\mathbf{p^i_{t-k}} = \mathbf{p^i_t} \mathbf{P_t} \mathbf{P^{-1}_{t-k}}
\end{equation}
Finally, each point \( \mathbf{p^i_{t-k}} \) is converted to voxel coordinates, which is the inverse process of Eqn.~\ref{eq:voxel2location}, and bilinear interpolation is used to sample features from corresponding BEV feature maps. The final cost volume feature map is \( \mathbf{F} \in \mathbb{R}^{H \times W \times Z \times (K \times N) \times C} \), where \( C \) represents the channels of the BEV features.
\subsection{Volume Features Refinement}
\begin{equation}
    \label{eq:refinement}
    \mathbf{V}_{\text{occ}} = \mathbf{V}_t \otimes \text{Sigmoid}(\text{Conv}(\mathbf{F}) )
\end{equation}
As described in the Eqn.~\ref{eq:refinement}, the constructed cost volume features \( \mathbf{F} \) are processed through a sequence of convolutional layers, yielding an output weight \(\mathbf{W} \in \mathbb{R}^{H \times W \times Z} \). A Sigmoid activation function then normalizes the output weights \( \mathbf{W} \) to the range [0,1]. These weights are directly supervised according to the voxel occupancy state: weights corresponding to occupied voxels are encouraged to reach 1, while those for unoccupied voxels are guided towards 0. The symbol \(\otimes\) in the Eqn.~\ref{eq:refinement} denotes element-wise product of the original volume features at timestamp $t$ with the learned weights \( \mathbf{W} \). This supervised learning approach produces an occupancy-aware volume feature map \( \mathbf{V}_{\text{occ}} \in \mathbb{R}^{H \times W \times Z \times C} \). The goal of the learned weights is to diminish the influence of voxel features in incorrectly activated regions due to depth ambiguity and augment the features of correctly identified voxels simultaneously. 
\subsection{Occupancy Decoder}
Upon the refinement of voxel features $\mathbf{V}_{\text{occ}}$, our model employs a series of deconvolution layers to transform it into occupancy features. The occupancy features are projected onto the output space, yielding $\mathbf{X} \in \mathbb{R}^{H \times W \times Z \times M}$. The projection is formulated to map the occupancy features into a discrete set of semantic class predictions for each voxel within the grid. Consequently, the final occupancy output $\mathbf{Y} \in \{ c_0, c_1, \ldots, c_M \}^{H \times W \times Z}$ is generated, where each voxel is assigned one of $M$ semantic labels, including an unoccupied state represented by $c_0$. This step effectively translates the enriched voxel features into a semantically segmented 3D occupancy map.
\subsection{Training Loss}
Our architecture integrates the CVT module and 3D occupancy prediction within an end-to-end trainable framework. We train the overall network with a multi-task loss as
\begin{equation}
\text{Loss} = \mathcal{L}_{\text{occ}} + \lambda \mathcal{L}_{\text{cvt}}
\end{equation}
where $\mathcal{L}_{\text{occ}}$ denotes the occupancy prediction loss, and $\mathcal{L}_{\text{cvt}}$ represents the cost volume temporal loss. The coefficient $\lambda$ serves as a balancing factor between these two loss components.

For the loss of occupancy prediction  \( \mathcal{L}_{\text{occ}} \), we adopt the cross-entropy loss, which is widely used in semantic occupancy prediction task. $\mathcal{L}_{\text{occ}} = \sum_c w_c \mathcal{L}(g_c, p_c)$, where $g_c$ and $p_c$ represent the ground truth label and the
prediction result for the $c$-th semantic class, respectively. Class-specific weighting $w_c$ mitigates class imbalance by inversely correlating with class frequency, as adopted from Occ3D~\cite{tian2023occ3dlargescale3doccupancy}.

To enhance the network's performance, the CVT module receives direct supervision. The loss attributed to the CVT module, $\mathcal{L}_{\text{cvt}}$, employs a binary cross-entropy framework as shown in Eqn.~\ref{eq:loss_cvt}:
\begin{equation}
\label{eq:loss_cvt}
\mathcal{L}_{\text{cvt}} = - \sum_{j=1}^{J} \hat{y}_{j,0} \log (1 - w_j) + \hat{y}_{j,1} \log w_j
\end{equation}
where $j$ indexes voxels in the grid, with $J = H \times W \times Z$. $w_j$ is the $j$-th element in the output weight $\mathbf{W} \in \mathbb{R}^{H\times W\times Z}$. Specifically, a binary label $\hat{y}_{j,0} = 1$ signifies the voxel's unoccupied status (i.e., its semantic classification is $c_0$), while $\hat{y}_{j,1} = 1$ indicates the voxel is occupied in ground truth. This direct supervision on the CVT module not only improves performance but also underpins the importance of accurately learning temporal and spatial features from the input data.
\section{Experiment}
\subsection{Dataset}
\noindent\textbf{Occ3D-Waymo.} We conduct experiments on the Occ3D-Waymo~\cite{tian2023occ3dlargescale3doccupancy} dataset, which provides dense semantic labels for 3D occupancy grids. Occ3D-Waymo consists of 798 training scenes and 202 validation scenes. On the Occ3D-Waymo dataset, the range for the $x$ and $y$ axis as $[-40m, 40m]$, and for the $z$ axis as $[-1m, 5.4m]$. The voxel grid size is $(0.4m, 0.4m, 0.4m)$, resulting in a resolution of $(200\times 200\times 16)$ for $(H, W, Z)$. while the semantic labels encompass 16 categories, including ``Free'' class. Besides, it also provides visibility masks for LiDAR and camera modality.

\noindent\textbf{Evaluation Metrics.} In our assessment of 3D semantic occupancy prediction capabilities, we utilize the IoU of each class to evaluate 3D semantic occupancy prediction. As Occ3D-Waymo provides camera visibility masks, only voxels within the camera's visible region are evaluated. The mean Intersection over Union (mIoU) is computed over 14 non-Free classes, excluding the ``Motorcycle'' class due to its insufficient voxel numbers. 
\subsection{Experimental Settings}
\noindent\textbf{Architecture Detail.} Our architecture leverages the same image backbone and image neck as Occ3D~\cite{tian2023occ3dlargescale3doccupancy} for image feature extraction. 
The transformation of image features to BEV space involves a four-layered view transformation layers, each layer comprising a sequence of cross-attention layers, normalization layers, feed-forward layers, and normalization layers. This BEV encoder generates a BEV feature representation $\mathbf{F^{\text{bev}}} \in \mathbb{R}^{200 \times 200 \times 256}$, which is further reshaped to volume features $\mathbf{V} \in \mathbb{R}^{200 \times 200 \times 16 \times 16}$. The selected frame configuration includes a time interval of $0.5s$ and a sequence of $7$ frames, covering a total time span of $3s$, which ensures a comprehensive temporal analysis. For the CVT module, by default, it samples $9$ points within each volume feature across $7$ frames to capture a broad spectrum of temporal and spatial variations. Consequently, this configuration enables the CVT module to generate a cost volume feature map $\mathbf{F} \in \mathbb{R}^{200 \times 200 \times 16 \times (7 \times 9) \times 16}$. The occupancy decoder utilizes a series of deconvolution layers to transform the volume features into occupancy features, which is then followed by two additional convolution blocks responsible for generating the final 3D semantic occupancy predictions.

\noindent\textbf{Training.} During training, we utilized the AdamW ~\cite{AdamW} optimizer for 8 epochs and full dataset each epoch. Training was conducted on 8 NVIDIA A100 GPUs with a batch size of 1 per GPU. Learning Rate Scheduler Policy is the cosine annealing schedule with an initial learning rate of $4\times 10^{-4}$. The dimensions for the input images were standardized to 960$\times$640 pixels. It is noteworthy that we abstained from employing any data augmentation techniques in order to ensure a fair and unbiased comparison among the models.
\subsection{Main Results}
As shown in Figure~\ref{fig:teaser}, We categorize previous temporal fusion methods into three types: (a) Attention-based methods, which adopt temporal self-attention to recurrently fuse the history BEV information; (b) WarpConcat-based methods, involving the alignment of BEV feature maps across different time steps; and (c) Plane-Sweep-based methods, which construct a cost volume for image pairs to utilize geometric constraints. Additionally, we introduce a novel approach, CVT-Occ, that utilizes geometric correspondences of 3D voxels over time to improve occupancy prediction accuracy. For comparative analysis, we extended some mainstream BEV models for the 3D semantic occupancy prediction task. We choose BEVFormer~\cite{li2022bevformer}, BEVFormer-WarpConcat~\cite{yang2022bevformerv2}, and SOLOFusion~\cite{solofusion} as representative methods for categories (a), (b), and (c), respectively. 

For both BEVFormer~\cite{li2022bevformer} and SOLOFusion~\cite{solofusion}, we have replaced their original transformer decoders for object detection with a 3D occupancy prediction decoder. Additionally, for fairness, we omit the depth supervision module when reproducing SOLOFusion~\cite{solofusion}. For BEVFormer-WarpConcat, we adopt a commonly used fusion method known as the Warp and Concatenate strategy, as seen in BEVFormerv2~\cite{yang2022bevformerv2} and PanoOcc~\cite{wang2023panooccunifiedoccupancyrepresentation}. This strategy involves 3D coordinate alignment in BEV space based on the transformation matrix between historical and current frames. Subsequently, it concatenates previous BEV features with current BEV features along the channel dimension and applies residual blocks for temporal fusion. We incorporate the residual block proposed by PanoOcc~\cite{wang2023panooccunifiedoccupancyrepresentation} to create the BEVFormer-WarpConcat method. In the ``BEVFormer-w/o TSA'' configuration, we maintain the original architecture of BEVFormer~\cite{li2022bevformer} but exclude the Temporal Self-Attention(TSA) layers, thereby eliminating temporal attention mechanisms. In Table \ref{tab:main_results}, we present the evaluation results for 3D occupancy prediction on the Occ3D-Waymo validation set. To ensure a fair comparison, all methods were implemented using the same network architecture and training strategy. 

\begin{table*}[t]
  \centering
\caption{\textbf{3D Occupancy Prediction Results on Occ3D-Waymo}. This comparison showcases the performance of various principal temporal fusion strategies alongside CVT-Occ, within the same architectural framework and identical frame specifications (number and interval), ensuring a direct and fair comparison across all evaluated methods. The leading performance is highlighted in \textbf{bold}.}
  \label{tab:main_results}
  \renewcommand\tabcolsep{0.5pt}
  \renewcommand{\arraystretch}{1.3}
  \tiny
  \begin{tabular}{l|c|*{15}{c}}
    \hline
    Method & \rotatebox{90}{\tiny mIoU} & \rotatebox{90}{\tiny Go} & \rotatebox{90}{\tiny Vehicle} & \rotatebox{90}{\tiny Pedestrian} & \rotatebox{90}{\tiny Sign} & \rotatebox{90}{\tiny Bicyclist} & \rotatebox{90}{\tiny Traffic Light} & \rotatebox{90}{\tiny Pole} & \rotatebox{90}{\tiny Cons. Cone} & \rotatebox{90}{\tiny Bicycle} & \rotatebox{90}{\tiny Building} & \rotatebox{90}{\tiny Vegetation} & \rotatebox{90}{\tiny Tree Trunk} & \rotatebox{90}{\tiny Road} & \rotatebox{90}{\tiny Walkable} 
    \\
    \hline
    BEVFormer-w/o TSA
    & 23.87 & \textbf{7.50} & 34.54 & 21.07 & 9.69 & \textbf{20.96} & 11.48 & 11.48 & 14.06 & 14.51 & 23.14 & 21.82 & 8.57 & 78.45 & 56.89 \\
    BEVFormer~\cite{li2022bevformer}  & 24.58 & 7.18 & 36.06 & 21.00 & 9.76 & 20.23 & 12.61 & 14.52 & 14.70 & 16.06 & 23.98 & 22.50 & 9.39 & 79.11 & 57.04  \\
    SOLOFusion~\cite{solofusion} & 24.73 & 4.97 & 32.45 & 18.28 & 10.33 & 17.14 & 8.07 & 17.83 & 16.23 & 19.3 & \textbf{31.49} & \textbf{28.98} & \textbf{16.93} & 70.95 & 53.28 \\
    BEVFormer-WarpConcat & 25.07 & 6.2 & 36.17 & 20.95 & 9.56 & 20.58 & \textbf{12.82} & 16.24 & 14.31 & 16.78 & 25.14 & 23.56 & 12.81 & 79.04 & 56.83  \\
    CVT-Occ(ours) & \textbf{27.37} & 7.44 & \textbf{41.0} & \textbf{23.93} & \textbf{11.92} & 20.81 & 12.07 & \textbf{18.03} & \textbf{16.88} & \textbf{21.37} & 29.4 & 27.42 & 14.67 & \textbf{79.12} & \textbf{59.09} \\
    \hline
  \end{tabular}
\end{table*}

Comparing ``BEVFormer-w/o TSA'' with temporal fusion methods in Table \ref{tab:main_results}, All temporal fusion methods demonstrate improved mIoU compared to ``BEVFormer-w/o TSA''. It becomes evident that temporal fusion plays a crucial role in the 3D occupancy prediction task. 

Maintaining consistent frame specifications (number and interval), we explore the temporal fusion ability of different mainstream temporal fusion methods. Our method achieves state-of-the-art performance in terms of mIoU and demonstrates significant superiority over other approaches across nearly all classes. As shown in Table \ref{tab:main_results}, CVT-Occ exhibits superior performance compared to representative models of existing temporal fusion methods (a), (b), and (c) in Figure \ref{fig:teaser}, achieving a notable 2.8\% mIoU improvement over the baseline BEVFormer~\cite{li2022bevformer}. Delving into the details, CVT-Occ shows improvements in IoU across almost all classes. Specifically, for 
``Vehicle'', ``Bicycle'', ``Building'', ``Vegetation'', and ``Tree Trunk'', our method achieves more than a 4\% increase in IoU compared to the baseline model BEVFormer~\cite{li2022bevformer}. This substantial enhancement across nearly all categories underscores that a comprehensive understanding of temporal geometry can lead to significant performance gains.
\subsection{Ablation Study}
\begin{figure}[t!]
  \begin{minipage}[h]{0.45\textwidth}
    \centering
    \renewcommand\tabcolsep{1.5pt}
    \renewcommand{\arraystretch}{1.3}
    \begin{tabular}[h]{l|ccc|c}
      \hline
    Exp & Frame & Interval & \begin{tabular}[m]{@{}c@{}}CVT \\supervision\end{tabular} & mIoU \\
    \hline
    (a) & 3 & 0.1s & \checkmark & 25.15 \\ \hline
    (b) & 3 & 0.3s & \checkmark & 25.86 \\ \hline
    (c) & 7 & 0.1s & \checkmark & 25.80 \\ \hline
    (d) & 7 & 0.3s & \checkmark & 26.46 \\ \hline
    (e) & 7 & 0.5s & \checkmark & \textbf{27.37} \\ \hline
    (f) & 7 & 0.5s &  & 26.81 \\
    \hline
    \end{tabular}
    \captionof{table}{
    \textbf{Ablation Experiments.} We evaluate CVT-Occ under different frame specifications and CVT supervision.
    }
    \label{tab:ablation}
  \end{minipage}
  \hfill
  \begin{minipage}[h]{0.5\textwidth}
    \centering
    \includegraphics[width=\linewidth]{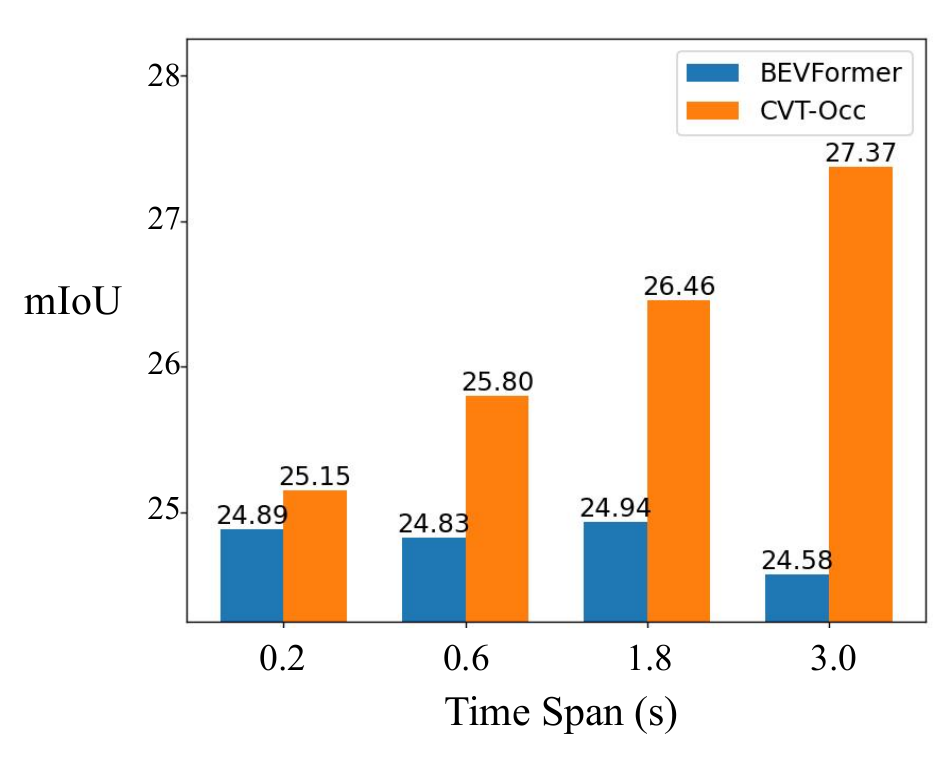}
    \caption{\textbf{Experiments of Different Time Spans.}}
    \label{fig:different_length}
  \end{minipage}
\end{figure}

\noindent\textbf{Time Span.} The time span is defined as the difference between the timestamp of the earliest historical frame and that of the current frame, which plays a crucial role in the temporal fusion analysis. We conduct experiments to evaluate our method across various configurations of total frame numbers and sampled frame intervals, as summarized in Table ~\ref{tab:ablation}. 

In experiments (a) and (c), we compared the impact of different total frame numbers. The results demonstrate that a longer sequence of historical frames leads to improved performance. Similarly, we observed that a larger time interval between adjacent frames also contributes to enhanced performance in experiments (c), (d), and (e). These consistent findings underscore the significance of a longer time span across the entire temporal sequence for better model performance. CVT-Occ relies on disparities between frames to mitigate depth ambiguity, where larger intervals and frame numbers contribute to an extended time span, amplifying the temporal parallax. This elucidates the observed increase in mIoU with a longer time span. 

In addition, comparing experiments (b) and (c), where the total time span remains the same $0.6s$, highlights the critical importance of the total time span even when the frame interval and number of historical frames varies between these two experiments. This finding further emphasizes that the impact of our proposed method arises from leveraging the parallax of the entire temporal sequence.

\noindent\textbf{Ability of Long Temporal Fusion.} Figure ~\ref{fig:different_length} illustrates the comparison between BEVFormer~\cite{li2022bevformer} and CVT-Occ across varying total time differences. CVT-Occ demonstrates improved performance with increasing temporal queue lengths, whereas BEVFormer struggles to effectively leverage information from long-range historical data. 

The limitation of BEVFormer's long temporal fusion capability has been highlighted in previous researches, such as UniFusion~\cite{qin2023unifusionunifiedmultiviewfusion} and VideoBEV~\cite{VideoBEV}. Based on results of Figure ~\ref{fig:different_length}, we propose that BEVFormer fails to benefit from longer time spans due to its recurrent temporal fusion process. This process restricts the direct access to information from distant historical frames, regardless of the temporal sequence length. In contrast, the cost volume feature $\mathbf{F}$ can directly integrate information from all historical frames. The CVT module utilizes parallax correlation explicitly to refine volume features, leading to a more robust understanding of temporal information. This design overcomes the shortcomings observed in BEVFormer~\cite{li2022bevformer}, resulting in enhanced long-term temporal fusion capability.

\noindent\textbf{CVT Supervision.} In our pursuit to improve 3D semantic occupancy prediction, we go beyond traditional loss functions by adding extra supervision to the CVT module output. Experiments (e) and (f) in Table~\ref{tab:ablation} shows the performance comparison of CVT-Occ with and without the CVT loss. Incorporating the CVT loss results in a significant increase in mIoU, demonstrating its effectiveness through experimental results. This enhancement validates our design approach, highlighting its ability to enhance the accuracy and reliability of semantic occupancy predictions in 3D. 

In conclusion, our ablation study confirms the critical importance of longer time spans and effective temporal fusion in enhancing CVT-Occ's performance for 3D semantic occupancy prediction. Additionally, integrating supplementary supervision further improves model accuracy, highlighting our approach's effectiveness in advancing temporal parallax-based scene understanding.

\subsection{Analysis of Factors}
\begin{table}[!t]
\centering
    \renewcommand\tabcolsep{2.0pt}
    \renewcommand{\arraystretch}{1.2}
    \caption{\textbf{Comparison Results under Different Conditions.} Here are three sets of experiments. (a) Binary Classification: this task focuses on geometric occupancy prediction, treating all object classes as a single ``Non-Free'' class. (b) BEV Range: Along x-axis, $[0, 20]$ means the range $[-20m, 0m]\cup [0, 20m]$ and $[20, 40]$ means the range $[-40m, -20m]\cup [20, -40m]$. (c) Ego Vehicle Speed: The 202 scenes for evaluation is partitioned into 2 parts based on the ego-vehicle speed. We evaluate all temporal fusion methods on them separately. }
    \label{tab:comparison}
    \begin{tabular}{l|c|cc|cc|cc}
      \hline
      Exp & mIoU & Non-Free & \hspace{0.5em}Free\hspace{0.5em} & \hspace{0.5em}$[0, 20]$ \hspace{0.5em}& \hspace{0.5em}$[20, 40]$\hspace{0.5em} & \hspace{0.5em}Fast \hspace{0.5em}& \hspace{0.5em}Slow \hspace{0.5em}\\
      \hline
      BEVFormer~\cite{li2022bevformer} & 24.58 & 55.74 & 86.54 & 27.09 & 21.24 & 24.73 & 24.22\\
      BEVFormer-WarpConcat  & 25.07 & 56.57& 86.96& 28.21 & 21.16 & 26.27 & 24.44 \\
      CVT-Occ  & 27.37 & 62.95& 90.46& 30.85 & 22.67 & 27.9 & 26.79\\
      \hline
    \end{tabular}
\end{table}

The core insight of our method lies in leveraging parallax between different frames to enhance the accuracy of 3D voxel depth queries. In this subsection, we conduct several evaluations under various conditions to validate this concept. These experiments include binary classification, different BEV ranges, and varying ego vehicle speeds.

\noindent\textbf{Binary Classification.}
Within the framework of 3D semantic occupancy prediction, we aggregate all 15 other object classes into a single ``Non-Free'' class. This adjustment removes the influence of semantic information, allowing us to focus purely on depth accuracy. We then compare our approach against leading temporal fusion methods. The results in Table ~\ref{tab:comparison} demonstrate that our method provides better predictions on object geometry. This outcome aligns closely with our initial design concept of leveraging parallax for refined depth prediction.

\noindent\textbf{BEV Range.} In the Occ3D-Waymo dataset, the BEV range spans from $-40m$ to $40m$ along the x-axis (i.e., the ego car's direction of movement). We divide this range into two parts: $[-20m, 20m]$ and $[-40m, 20m] \cup [20m, 40m]$. Table ~\ref{tab:comparison} exhibits the evaluation results. All methods perform better in the near region, which is expected since objects closer to the vehicle are more likely to appear in the overlap region of history and current frames. However, our method exhibits significantly greater improvement in the near region(+3.76) compared to the far region(+1.43) compared to BEVFormer~\cite{li2022bevformer}. The larger parallax of nearer objects between different frames contributes to our method's superior prediction capability.

\noindent\textbf{Ego Vehicle Speed.} Moreover, the speed of the vehicle also plays a crucial role in determining the magnitude of parallax. We further divide the dataset into two roughly equal parts based on the speed of ego vehicle. Evaluating CVT-Occ separately on these parts yields the results shown in Table ~\ref{tab:comparison}. Regardless of whether BEVFormer~\cite{li2022bevformer} or CVT-Occ, both perform better in fast-moving scenarios. Our method demonstrates much greater improvement under the fast condition(+3.17) than slow one(+2.57) compared to the baseline BEVFormer. This can be attributed to the larger parallax between history and current frames when the ego car is moving rapidly.

To sum up, these results underscore the significantly enhanced performance of CVT-Occ compared to the baseline model BEVFormer~\cite{li2022bevformer} under a wide range of conditions. The evaluations demonstrate that our method can refine the depth estimation of 3D voxel queries more accurately, with a notable advantage in the near region and fast scenes due to the greater parallax effect.
\subsection{Visualization Results}
\begin{figure*}[t!]
  \centering
   \includegraphics[width=\linewidth]{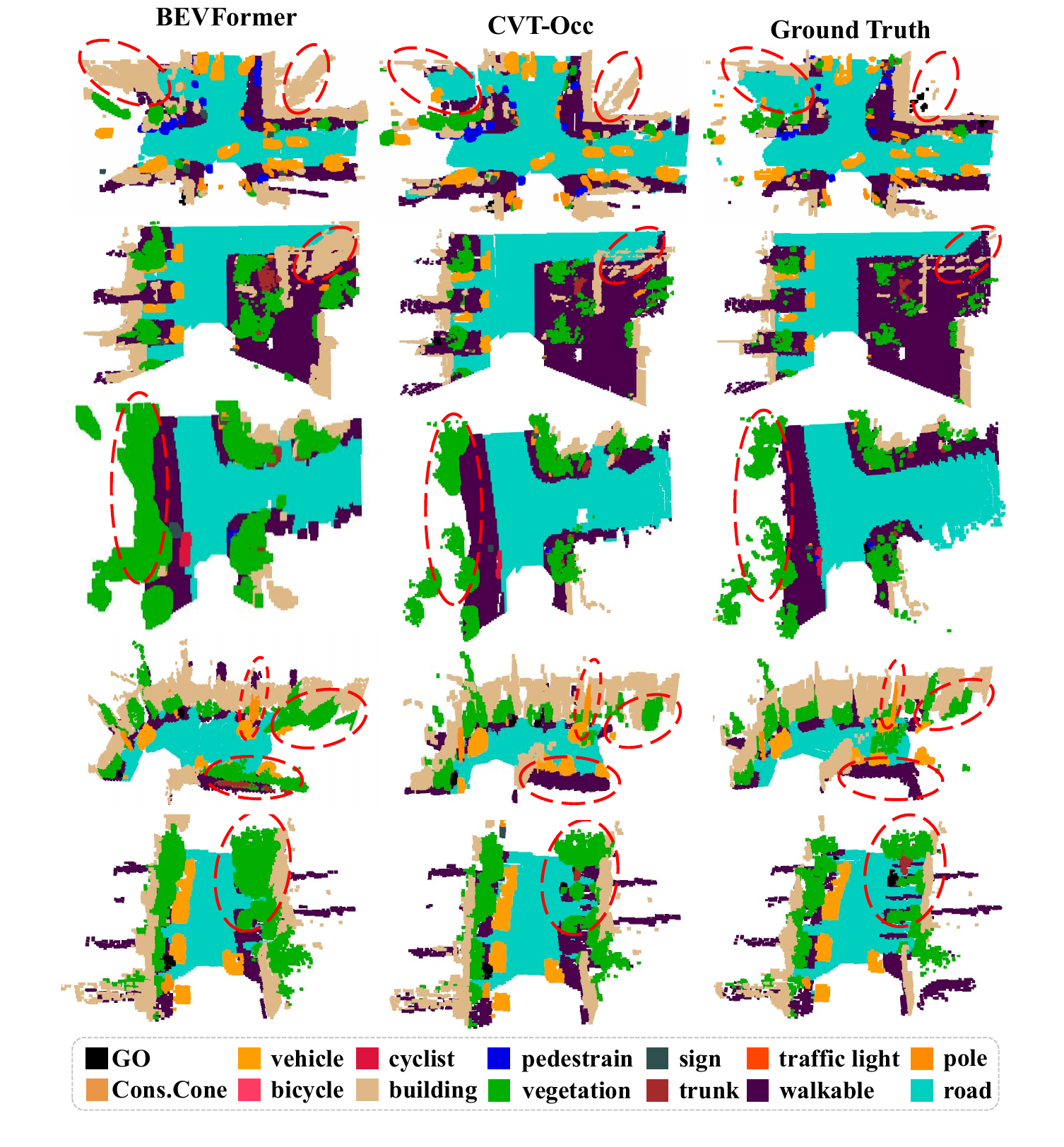}
   \caption{\textbf{Qualitative Results.} CVT-Occ exhibits superior performance in predicting the occupancy of vegetation and buildings. 
   }
   \label{fig:visualizations}
\end{figure*}
Figure \ref{fig:visualizations} offers a qualitative comparison of CVT-Occ against the baseline method BEVFormer~\cite{li2022bevformer} on the Occ3D-Waymo dataset. To ensure clarity in visualization, a dataset-specific mask is applied to exclude regions not within the camera's field of view. These visual representations clearly illustrate the superiority of CVT-Occ over competing approaches. BEVFormer encounters difficulties with depth perception, leading to issues such as atmospheric distortion and blurring of objects. CVT-Occ marks a remarkable improvements in the accurate depiction of buildings and vegetation, thus providing a more precise and reliable interpretation of the environment.
\section{Discussion and Conclusion}
We have presented CVT-Occ, a novel approach that significantly enhances the accuracy of 3D occupancy predictions by leveraging temporal fusion and geometric correspondence across time. Unlike traditional methods that rely on monocular or stereo vision, our method utilizes multi-view temporal fusion, effectively incorporating historical observations to exploit the parallax effect and improve depth estimation.
Our innovative approach, which constructs a cost volume feature map by sampling and integrating features across temporal frames, demonstrates superior performance on the Occ3D-Waymo dataset, surpassing current state-of-the-art models in terms of accuracy while maintaining a low computational overhead. 
CVT-Occ also opens avenues for future research on temporal fusion for 3D perception, \eg, applying the cost volume module in other tasks like 3D reconstruction, and other applications like robotics and virtual reality.  

\noindent\textbf{Acknowledgments.} This work is supported by the National Key R\&D Program of China (2022ZD0161700).


%
%
\bibliographystyle{splncs04}
\bibliography{main}

\end{document}